\def\year{2019}\relax
\documentclass[letterpaper]{article} 
\usepackage{aaai19}  
\usepackage{times}  
\usepackage{helvet}  
\usepackage{courier}  
\usepackage{url}  
\usepackage{graphicx}  
\usepackage{url}
\usepackage{booktabs}
\usepackage{multirow}
\usepackage{colortbl}
\usepackage{latexsym}
\usepackage{amsmath}
\usepackage{amssymb}
\begin{document}
	
	\title{Gaussian Word Embedding with a Wasserstein Distance Loss}
	
	\author{
		Chi Sun, Hang Yan, Xipeng Qiu and Xuanjing Huang\\
		Shanghai Key Laboratory of Intelligent Information Processing, Fudan University\\
		School of Computer Science, Fudan University\\
		825 Zhangheng Road, Shanghai, China\\
		\{sunc17,hyan,xpqiu,xjhuang\}@fudan.edu.cn
	}
	
	\maketitle
	\begin{abstract}
		Compared with word embedding based on point representation, distribution-based word embedding shows more flexibility in expressing uncertainty and therefore embeds richer semantic information when representing words. The Wasserstein distance provides a natural notion of dissimilarity with probability measures and has a closed-form solution when measuring the distance between two Gaussian distributions. Therefore, with the aim of representing words in a highly efficient way, we propose to operate a Gaussian word embedding model with a loss function based on the Wasserstein distance. Also, external information from ConceptNet will be used to semi-supervise the results of the Gaussian word embedding. Thirteen datasets from the word similarity task, together with one from the word entailment task, and six datasets from the downstream document classification task will be evaluated in this paper to test our hypothesis.
	\end{abstract}
	
	\section{Introduction}
	To model language, we need to represent words. We can use a one-hot method \cite{baeza1999modern} to represent words, but the dimensions of the words represented by this method are equal to the vocabulary size, which will easily lead to the \emph{curse of dimensionality}. Also, it does not contain the semantic information and cannot measure the dissimilarity between words. To overcome the shortcomings of the one-hot method, \citeauthor{bengio2003neural} \shortcite{bengio2003neural} propose a distributed representation of words, using neural networks to train language models. A commonly used distributed word representation is the word2vec model \cite{mikolov2013efficient} which is based on the distributional hypothesis: words that occur in the same contexts tend to have similar meanings \cite{harris1954distributional}. 
	
	Instead of being represented by a deterministic point vector, words can also be represented by a whole probability distribution. \citeauthor{vilnis2014word} \shortcite{vilnis2014word} model each word by a Gaussian distribution, and learn its mean and covariance matrix from data. Their model has many advantages such as providing more similarity measures, better grasping uncertainty, and measuring asymmetric relations like entailment in a more scientific way. For example, compared to cosine similarity and Euclidean distance represented by the points, using the distribution-based representation can extend the definition of indicators that measure the similarity of two words, such as KL divergence, Wasserstein distance, etc. However, \citeauthor{vilnis2014word} \shortcite{vilnis2014word} use a loss function based on KL divergence. KL divergence is ill-defined for two very similar distributions, and it is not sensitive to the distance between two distant distributions. It seems to be inconclusive to use KL divergence as the only energy function to train an entire corpus.
	
	In this paper, we propose to train the Gaussian word embeddings with a Wasserstein distance loss. Wasserstein distance \cite{olkin1982distance}, also known as the earth mover's distance, has been used in geometry \cite{ni2009local} and image processing \cite{gu2013variational} for a long time. \citeauthor{frogner2015learning} \shortcite{frogner2015learning} first proposed supervised learning using Wasserstein distance as a loss function. Wasserstein distance does not have a closed-form solution in most cases. It usually uses an approximate criterion to calculate the distance, so its calculation speed is slow. Fortunately, when concerning two multidimensional Gaussian distributions, Wasserstein distance has a closed-form solution, so it is convenient to be calculated as a loss function of the Gaussian word embeddings.
	
	The traditional method of training word embeddings is unsupervised learning. To improve the quality of word vectors, external information is often used to conduct semi-supervised training. \citeauthor{wang2014knowledge} \shortcite{wang2014knowledge} use knowledge mapping information and \cite{tissier2017dict2vec} introduce dictionary information to train the word vectors in order to achieve significant results. \cite{faruqui2014retrofitting} introduce a graph-based retrofitting method where they post-process learned vectors concerning semantic relationships extracted from additional lexical resources. The methods above are deemed as optimizations for point vector word embeddings. For Gaussian word embeddings, we introduce the external information from ConceptNet~\cite{speer2012representing}, treat the information as another form of context and define the energy function based on alternative metrics for semi-supervised training. In particular, we use KL divergence for the `IsA' relation and use Wasserstein distance for all the relations mentioned in ConceptNet. The asymmetry of the relationship between hypernym and hyponym makes the energy function based on KL divergence reasonable.
	
	It turns out that our model outperforms other alternative methods in the word similarity task. And we also prove the lifting effect of the semi-supervised method through the word entailment dataset. In the downstream document classification task, our model also achieves improvements over other alternative models.
	
	\section{Related Work}
	\subsection{Point-based word embeddings}
	In the original model from \cite{bengio2003neural}, a feedforward neural network is used to predict missing words in sentences. The trained hidden layer parameters are used as word embeddings. There are also improvements in replacing fully-connected networks with recurrent neural networks \cite{mikolov2010recurrent,mikolov2011extensions}. \citeauthor{collobert2008unified} \shortcite{collobert2008unified} first used a window approach to feed a neural network and learn word embeddings. And the most widely used algorithms are the continuous bag of words and skip-gram models \cite{mikolov2013efficient,mikolov2013distributed}, which use a series of optimization methods such as negative sampling \cite{gutmann2012noise} and hierarchical softmax \cite{mnih2009scalable}.
	
	Another way to learn word embeddings is through factorization of word co-occurrence matrices such as GloVe embeddings \cite{pennington2014glove}. This method of matrix factorization has been shown to be intrinsically linked to skip-gram and negative sampling \cite{levy2014neural}.
	
	\subsection{Ways to improve standard ``point" embeddings}
	Combining probabilistic models is a new idea for training word embeddings. \citeauthor{liu2015topical} \shortcite{liu2015topical} propose topical word embeddings and combine a probabilistic topic model with word embeddings. \citeauthor{shi2017jointly} \shortcite{shi2017jointly} further explore and propose a joint training topic model and word embedding method. Expanding on the basis of the words, for different distributed data, \citeauthor{rudolph2016exponential} \shortcite{rudolph2016exponential} propose an embedded model of the exponential distribution family, defining the embedding from a more unified perspective. \citeauthor{nickel2017poincare} \shortcite{nickel2017poincare} present an embedding model that replaces standard euclidean space with hyperbolic space. \citeauthor{vilnis2014word} \shortcite{vilnis2014word} propose a Gaussian distribution to model each word which uses an energy function based on KL divergence.
	
	\subsection{The application of Wasserstein distance}
	In addition to KL divergence, Wasserstein distance can also be used to measure the relationship between two distributions. Even if the support sets of the two distributions do not overlap or overlap very little, the distance between the two distributions can still be reflected by Wasserstein distance. Wasserstein distance is a well-characterized measure \cite{olkin1982distance}. It is commonly used in the field of image processing \cite{gu2013variational}. The Wasserstein GAN leverages Wasserstein distance to produce a value function which has better theoretical properties than the original \cite{arjovsky2017wasserstein}. \citeauthor{tolstikhin2017wasserstein} \shortcite{tolstikhin2017wasserstein} propose the Wasserstein Auto-Encoder for building a generative model of data distribution. \citeauthor{frogner2015learning} \shortcite{frogner2015learning} first used Wasserstein distance as the loss function of a neural network. Recently, \cite{Singh2018vector} propose a unified framework for building unsupervised representations of individual objects or entities with Wasserstein distance. The focus of \citeauthor{Singh2018vector}'s work and our work is not the same, and we are more concerned with the learning and optimization of word-level Gaussian embeddings.
	
	\subsection{Learning with external information}
	Although the word embedding models mentioned above are effective, they suffer from a classic drawback of unsupervised learning: the lack of supervision between a word and those appearing in the associated contexts. Recent approaches have proposed to tackle this issue using external information. \citeauthor{wang2014knowledge} \shortcite{wang2014knowledge} use knowledge-mapping information and \cite{tissier2017dict2vec} introduce dictionary information to train the word vectors. Similarities derived from such resources are part of the objective function during the learning phase \cite{yu2014improving,kiela2015specializing} or used in a retrofitting scheme \cite{faruqui2014retrofitting}.
	
	In this paper, we use the loss function based on Wasserstein distance to train Gaussian word embeddings. Also, we incorporate new relations from ConceptNet and use KL divergence to train the `IsA' relation and Wasserstein distance to train all the relations in ConceptNet to improve the Gaussian embedding for different tasks.
	
	\section{Methodology}
	In this section, we introduce the Wasserstein Distance Gaussian (WDG) model for word representation. We take advantage of the good metric nature of Wasserstein distance and construct a loss function based on it. Regarding data processing, we use sub-sampling and randomization methods to improve the model performance.
	
	\subsection{Word Representation}
	We denote each word $w$ in the vocabulary with a D-dimensional Gaussian distribution. The probability density function of the Gaussian distribution of the word is as follows:
	\[
	\begin{split}
	f_{w}(\vec{x}) &= \mathcal N(\vec{x};\vec{\mu}_{w},\Sigma_{w}) \\
	& =\frac{1}{\sqrt{(2\pi)^D|\Sigma_{w}|}}e^{-\frac{1}{2}(\vec{x}-\vec{\mu_{w}})^\mathsf{T}\Sigma_{w}^{-1}(\vec{x}-\vec{\mu_{w}})}.
	\end{split}
	\]
	
	The mean vector $\mu_{w}$ represents the center point of the word $w$ in the feature space. We assume that the different dimensions of the multidimensional Gaussian distribution are independent of each other. Under this assumption, the covariance matrix $\Sigma$ is a diagonal matrix. The numerical magnitude of the determinant of the covariance matrix $\Sigma_{w}$ represents the size of the semantic range of the word $w$. For example, the semantic range of the word ``mammal" is larger than the word ``cat", and we hope that the trained covariance matrix can also satisfy this property.
	
	\subsection{Wasserstein distance}
	The ${p^{\text{th}}}$ Wasserstein distance between two probability measures $\mu$  and $\nu$  in the probability space $M$ is defined as follows:
	$$W_{p}(\mu,\nu):=\left(\displaystyle \inf \limits_{\gamma \in \Gamma (\mu,\nu)}	\int_{M \times M} d(x,y)^{p}d\gamma(x,y)\right)^{\frac{1}{p}},$$
	where $\Gamma$ is a joint distribution over $M \times M$, and must satisfy both $\mu$ and $\nu$. The metric $d$ can be any distance on probability space $M$, such as Euclidean distance, $L_{1}$ distance, etc.
	
	It is easy to verify that Wasserstein distance satisfies positive definiteness, symmetry, and triangular inequality. It is a well-defined measure.
	
	\subsection{Kullback-Leibler divergence}
	For distributions $P$ and $Q$ of a continuous random variable, the Kullback-Leibler divergence is defined to be the integral:
	$${\displaystyle D_{\mathrm {KL} }(P\|Q)=\int _{-\infty }^{\infty }p(x)\,\log {\frac {p(x)}{q(x)}}\,dx}.$$
	KL divergence has some drawbacks when it is used as an energy function to training Gaussian word embeddings. We will give a specific explanation in the next subsection.
	
	\subsection{Comparison between Wasserstein distance and KL divergence}
	The most significant difference between Wasserstein distance and KL divergence is symmetry. The former considers the metric nature of the distribution and $D_{\mathrm {KL} }(P\|Q)$ only pays attention to the magnitude of $Q$ in the place where $P$ has a large value. The KL divergence is ill-defined for two very similar distributions:
	$$P=\mathcal N(0, \epsilon^3) ,Q=\mathcal N(\epsilon, \epsilon^3).$$
	We can calculate $D_{\mathrm {KL}}(P\|Q)=\frac{1}{2\epsilon}
	$, which means that when $\epsilon \rightarrow 0$, there should be $P \rightarrow Q$, but the KL divergence is divergent.
	
	Under Gaussian distribution conditions, Wasserstein distance has an analytical solution. We can calculate
	$W_{2}(P,Q)=\epsilon$, which means that when $\epsilon \rightarrow 0$, we can conclude that $W_{2}(P,Q)\rightarrow 0$. The result is reasonable.
	
	On the other hand, the advantage of Wasserstein distance is that it can well grasp the distance between two distant distributions. Even if the support sets of the two distributions do not overlap or overlap very little, Wasserstein distance can still reflect the distance between them. However, KL divergence is not sensitive to the distance between such two distributions. 
		
	However, the disadvantages above do not mean that KL divergence is useless. The characteristics of KL divergence make it more suitable for measuring the entailment relationship. Therefore, we use KL divergence in the energy function of the hypernym.
	
	\subsection{Objective function of WDG model}
	The Gaussian word embedding is based on the skip-gram framework. We use the energy-based loss function. The samples in WDG model contain the word $w$, the context $c$ of the word $w$, and $k$ negative examples $n_1,...,n_k$. The word $w$ is a word selected from a sentence in the corpus, and the context $c$ is a nearby word within a window size $l$. The negative sample $n_i$ is a word that is randomly sampled at a certain frequency in the vocabulary. The loss function of WDG model is as follows:
	$$L=\log \sigma(E(w,c)) +\sum_{i=1}^{k} \log \sigma(-E(w,n_i)),$$
	where $\sigma$ is the sigmoid function,  $E(w,c)$ is the energy function that we will introduce next.
	
	Our goal is to maximize $E(w,c)$ and minimize $E(w,n)$. We choose Wasserstein distance as the energy function. We know that the more similar $w$ and $c$ are, the smaller their Wasserstein distance is, and we want to maximize the loss function, so we need to add a minus sign before the formula: $$E(w,c)=-W_{2}(f_w,f_c)+b,$$
	where $W_{2}(f_w,f_c)$ is Wasserstein distance, $b$ is a bias due to the following reason:
	
	Note that the original meaning of the $\sigma$ is a conditional probability, so we need to limit its result to 0 to 1. Since Wasserstein distance and KL divergence are all positive, if they are used as an energy function without modification, the sigmoid function is always greater than 0.5, so we need to add a bias term $b$ so that the value of the energy function is in a suitable range.
	
	In general, it is difficult to find analytical solutions for Wasserstein distance. Fortunately, the Wasserstein distances between two Gaussian distributions are analytically resolved, which facilitates the training of Gaussian word embeddings.
	
	Assume that words $w_1$ and $w_2$ satisfy D-dimensional Gaussian distributions $\mathcal N_1(\vec{\mu_1},\Sigma_1)$ and $\mathcal N_2(\vec{\mu_2},\Sigma_2)$, respectively. Then the formula for Wasserstein distance is as follows:
	\[
	\begin{split}
	W_{2}(f_{w1},f_{w2})^2&=\|\vec{\mu_1}-\vec{\mu_2}\|_2^2+\\ &tr(\Sigma_1+
	\Sigma_2-2(\Sigma_2^\frac{1}{2}\Sigma_1\Sigma_2^\frac{1}{2})^\frac{1}{2}).
	\end{split}
	\]
	
	\subsection{Word sampling and batch shuffled}
	In a corpus, the semantic information contained in high-frequency words is often inferior to low-frequency words. For example, words such as ``a", ``the", ``it" are not as meaningful as ``plane", ``king", ``cat". In order to reduce the influence of high-frequency words, we use the sub-sampling method in word2vec and discard word $w$ with the probability of $P(w)=1-\sqrt{\frac{t}{f(w)}}$ when generating the batch, where $f(w)$ is the frequency of word $w$ and $t$ is a chosen threshold, typically around $10^{-5}$.
	
	For negative sampling, we use random sampling with the probability of $P(w)\propto U(w)^{\frac{3}{4}}$, where $U(w)$ is the unigram distribution, which is the frequency of single words appearing in the corpus. This method also plays a role
	in reducing the frequency of occurrence of high-frequency words.
	
	In cases where the data is affected by the order, disrupting the order of the data ensures that the gradient estimates are independent of each other. When skip-gram generates a batch, a central word $w$ will form pairs with several context words $c_1,...c_k$. We generate the word pairs $(w, c_i)$ in advance and then scramble them. This randomization method has achieved better results.
	
	\section{Incorporating new relations}
	
	To overcome the lack of supervision of unsupervised learning, we try to add external information from ConceptNet based on WGD model for semi-supervised learning and propose the Wasserstein Distance Gaussian embedding with external information (WDG-ei) model. We can flexibly select different energy functions for training according to the type of external information.
	
	\subsection{ConceptNet}
	
	ConceptNet\footnote{http://conceptnet.io/} \cite{speer2012representing} is a semantic network that contains a lot of information that the computer should know about the world. This information helps the computer to do a better search, answer questions, and understand human intent. It consists of nodes that represent concepts that are expressed in natural language words or phrases, and in which the relationships of these concepts are marked.
	
	The interlingual relations in ConceptNet are `IsA', `UsedFor', `CapableOf', etc.
	
	\subsection{Objective function of WDG-ei model}
	We define two parts of the loss function for WDG-ei model. Some of them are the same as WDG model. There is a word $w$, a context $c$, and $k$ negative examples $n_{11},...,n_{1k}$. The loss function of this part is as follows:
	$$L_{1}=\log \sigma(E_1(w,c)) +\sum_{i=1}^{k} \log \sigma(-E_1(w,n_{1i})).$$
	
	The other part of the loss function has a similar form. The difference is that the context $c$ is replaced by the external information $e$. The external information $e$ is randomly selected from ConceptNet. If the word $w$ does not have a related word in ConceptNet, $e$ will be set to a fixed special character. In addition, the negative sample $n_{2i}$ will be resampled. The loss function in this part is as follows:
	$$L_2=\log \sigma(E_2(w,e)) +\sum_{i=1}^{k} \log \sigma(-E_2(w,n_{2i})).$$
	
	The final loss function of the WDG-ei model is:
	$$L(w,c,e,n_{1},n_{2})=L_1(w,c,n_{1})+\alpha \cdot L_2(w,e,n_{2}),$$
	where $\alpha$ is a coefficient used to balance the effect of $E_1(w,c)$ and $E_2(w,e)$ on the global loss function.
	
	In the WDG-ei model, $E_1(w,c)$ is fixed:  $E_1(w,c)=-W_{2}(f_w,f_c)+b_1$. For the second energy function $E_2(w,e)$, we have more abundant choices, and we can choose it freely according to the type of external information. For example, Wasserstein distance can also be selected for the synonyms: $E_2(w,e)=-W_{2}(f_w,f_c)+b_2$, and KL divergence can be selected for the hyponyms: $E_2(w,e)=-D_{\mathrm {KL}}(f_w\|f_h)+b_2$.
	
	We select $E_2(w,e)=-W_{2}(f_w,f_c)+b_2$ by default and introduce all the relationships in ConceptNet for training, and the model is named as Wasserstein Distance Gaussian embedding with external information (WDG-ei) model.
	
	\subsection{Non-symmetric relation}
	Synonyms express the symmetric relationship, while hyponyms reflect the asymmetric relationship. The asymmetric relationship is more suitable for training with an asymmetric energy function, such as KL divergence.
	
	In order to make the trained word embeddings better embody the entailment relationship between words, we propose the WDG-ei(IsA) model, select $E_2(w,e)=-D_{\mathrm {KL}}(f_w\|f_h)+b_2$, and only introduce the `IsA' relation in ConceptNet for training.
	
	The formula for KL divergence of two words is as follows:
	\[
	\begin{split}
	D_{\mathrm {KL}}(f_{w1}\|f_{w2})&=\frac{1}{2}[log\frac{|\Sigma_2|}{|\Sigma_1|}-D+tr(\Sigma_2^{-1}\Sigma_1)+\\&(\vec{\mu_2}-\vec{\mu_1})^\mathsf{T}\Sigma_2^{-1}(\vec{\mu_2}-\vec{\mu_1})].
	\end{split}
	\]

	\begin{table*}[t!]
		\centering
		\begin{tabular}{c|c c c c|c c c c}
			\toprule
			Dataset & SG & W2G & WDG & WDG-ei & SG & W2G & WDG & WDG-ei \\
			\hline
			dim & \multicolumn{4}{ c |}{50} & \multicolumn{4}{c}{100} \\
			\hline
			MC-30 & 70.92 & 80.09 & 77.43 & \textbf{83.14} & 70.28 & 76.60 & 77.76 & \textbf{83.54} \\
			MEN-3k & 65.05 & 66.14 & 65.43 & \textbf{67.40} & 68.20 & 69.66 & 72.66 & \textbf{74.17} \\
			MTurk-287 & 64.58 & \textbf{65.35} & 58.90 & 63.23 & 66.88 & \textbf{66.98} & 65.44 & 66.79 \\
			MTurk-771 & 57.16 & 56.80 & 57.39 & \textbf{61.65} & 59.70 & 57.85 & 59.92 & \textbf{69.28} \\
			RG-65 & 70.98 & 73.94 & 74.53 & \textbf{78.45} & 74.06 & 75.28 & 75.70 & \textbf{85.62} \\
			RW & 35.09 & 38.54 & 40.71 & \textbf{42.63} & 39.15 & 39.99 & 41.34 & \textbf{44.11} \\
			SIMLEX-999 & 26.66 & 24.76 & 31.03 & \textbf{37.24} & 31.14 & 27.24 & 34.28 & \textbf{51.96} \\
			SimVerb-3500 & 17.07 & 14.91 & 19.93 & \textbf{27.14} & 19.78 & 16.21 & 22.97 & \textbf{44.52} \\
			VERB-143 & 28.04 & 28.04 & \textbf{31.43} & 24.65 & 31.22 & 30.04 & 32.79 & \textbf{40.88} \\
			WordSim-353-ALL & 60.64 & 61.36 & 63.12 & \textbf{69.17} & 63.43 & 64.21 & 68.18 & \textbf{71.05} \\
			WordSim-353-REL & 52.13 & 53.65 & 57.00 & \textbf{62.63} & 54.32 & 56.68 & 61.50 & \textbf{62.90} \\
			WordSim-353-SIM & 69.71 & 70.91 & 71.08 & \textbf{76.00} & 72.50 & 71.33 & 75.01 & \textbf{80.07} \\
			YP-130 & 22.69 & 35.98 & 41.50 & \textbf{51.21} & 28.78 & 36.46 & 48.13 & \textbf{64.82} \\
			\midrule
			Avg & 49.29 & 51.57 & 53.04 & \textbf{57.27} & 52.26 & 52.96 & 56.60 & \textbf{64.60} \\
			$\Delta$ & - & +2.28 & +3.75 & \textbf{+7.98} & - & +0.70 & +4.34 & \textbf{+12.34} \\ \bottomrule
		\end{tabular}
		\caption{Spearman correlation for word similarity datasets. The models SG, W2G, WDG, and WDG-ei represent word2vec skip-gram model \cite{mikolov2013efficient}, Gaussian embedding model \cite{vilnis2014word}, Wasserstein Distance Gaussian model, and Wasserstein Distance Gaussian model with external information, respectively. We uniformly use cosine similarity to compare word similarities. For different models of the same dimension, we bolded the best performing score.
		}
	\end{table*}
	
	\section{Experiments}
	
	\subsection{Preparation for training data}
	To learn the model parameters, we select the English dump from Wikipedia updated on March 1, 2018\footnote{https://dumps.wikimedia.org/enwiki/20180301/}. We clean the original data, remove non-English words and special symbols, convert uppercase letters to lowercase, and obtain the final corpus containing 2.4 billion tokens. We discard words that appear less than 100 times and obtain a vocabulary of size 302,570. In this process, we count the frequency of each word to prepare for the subsequent sampling work.
	
	The external information of the WDG-ei model is the relations `DefinedAs', `InstanceOf', `SimilarTo', `Synonym', `FormOf' and `IsA' in ConceptNet. We preprocess the data and discard words that are not in our vocabulary.
	
	\subsection{Hyperparameters}
	We assume that the word is a multidimensional Gaussian distribution. The dimensions are independent of each other, and the covariance matrix is diagonal. The uncertainty measured by the covariance matrix does not require a vast number of hyperparameters. We have found that training a spherical covariance matrix is much better than training diagonally. It can be considered that this assumption is a regularization, so in the experiment, we choose a spherical covariance matrix model for training.
	
	We test the 50-dimensional and 100-dimensional training results. The number of training iterations is 5, and the selected window size is 5. We set an initial learning rate of 0.025, and the learning rate will gradually decline as the training progresses. The subsampling parameter $t$ to discard high-frequency words is set to $10^{-5}$, and the number of negative samples per word is set to 5.
	
	\subsection{Word similarity evaluation}
	We use the traditional method of evaluating the similarity of words. The model is measured by calculating the Spearman’s rank correlation coefficient \cite{spearman1904proof} between the human similarity evaluation and the model similarity scores of the word pairs. The Spearman correlation is a rank-based correlation measure that assesses how well the scores describe the true labels.
	
	We use data sets MC-30 \cite{miller1991contextual}, MEN \cite{bruni2014multimodal}, MTurk-287 \cite{radinsky2011word}, MTurk-771 \cite{halawi2012large}, RG-65 \cite{rubenstein1965contextual}, RW \cite{luong2013better}, SIMLEX-999 \cite{hill2015simlex}, SimVerb-3500 \cite{gerz2016simverb}, VERB-143 \cite{baker2014unsupervised}, WordSim-353 \cite{finkelstein2001placing}, YP-130 \cite{yang2006verb}.
	
	We train word2vec skip-gram, word2Gaussian, WDG, and WDG-ei model. All models use the same corpus and the same hyperparameters during training, so the vocabularies of all models are the same. To reduce the uncertainty caused by random initialization, we run each experiment 3 times and average the results. We use a slightly modified version of the toolkit by \cite{faruqui-2014:SystemDemo}  to analyze word similarity. The experimental results are reported in Table 1.
	
	We find that on most datasets, WDG model performs better than the skip-gram model and word2Gaussian model. Also, we are pleasantly surprised to find that WDG-ei model which introduces ConceptNet's external information for training has dramatically improved the performance of similarity datasets, and even improve over $20\%$ in some data sets such as SimVerb-3500 and YP-130.
	
    We can choose more abundant energy functions for WDG-ei model when introducing external information. The energy function based on Wasserstein distance can be used when adding the synonym information, and the KL-based energy function can be used when introducing the hyponym information. We will report the performance of the entailment dataset when the WDG-ei model introduces the `IsA' relation in ConceptNet in the next subsection.
	
	\subsection{Word entailment evaluation}
	We evaluate our model on the word entailment dataset from \cite{baroni2012entailment}. The lexical entailment between words is denoted by $w_1\vDash w_2$ , meaning that all instances of $w_1$ are $w_2$. All words in the data set are nouns, with 1380 pairs of positive cases, such as $highway\vDash road$, and 1381 negative cases, such as  $highway\nvDash machine$.
	
	There are several point-embedding-based methods  \cite{nickel2017poincare,henderson2016vector,weeds2014learning,vulic2017specialising} and Gaussian-embedding-based methods\cite{vilnis2014word,athiwaratkun2017multimodal} to detect hypernymy. And the Gaussian embeddings have been proven powerful as they can capture uncertainty about the word. 
	
	We will select appropriate metrics to evaluate the performance of our model. It is worth noting that the entailment relationship is asymmetric. Measuring the asymmetric relationship with a symmetric metric such as cosine similarity is ambivalent because exchanging two words in a word pair makes a positive case negative, such as $horse\vDash mammal$ and $mammal\nvDash horse$. However, evaluating these word pairs above with cosine similarity results in the same score. Fortunately, we have a suitable asymmetric metric, KL divergence, to measure the entailment. We generate the KL divergence of word pairs as a score and select the best threshold to distinguish between positive and negative cases. We calculate the best F1 score and the best average precision and report the results in Table 2.
	
	\begin{table}[h!]
		\centering
		\begin{tabular}{c|c|c}
			\toprule
			Model & Best F1 & Best AP \\
			\hline
			W2G & 75.1 & 71.7 \\
			WDG & 75.7 & 72.7 \\
			WDG-ei(all) & 78.9 & 75.7 \\
			WDG-ei(IsA) & 79.6 & 76.2 \\
			\bottomrule
		\end{tabular}
		\caption{The best F1 scores and the best average precision values for entailment datasets. The W2G and WDG represent the Gaussion embedding \cite{vilnis2014word} and the Gaussian Wasserstein distance model, respectively. The WDG-ei(all) denotes that all relations in ConceptNet are selected, and the WDG-ei(IsA) denotes that only the ConceptNet's `IsA' relation is used as external information. The model uses 100-dimensional word embeddings with window size 5.
		}
	\end{table}	
	
	We find that the result of WDG-ei(all) model with all the external information in ConceptNet is much better than word2Gaussian model and WDG model. WDG-ei(IsA) model, which is targeted to add only `IsA' relation in ConceptNet, achieves the most excellent results on the entailment dataset.
	
	That is to say, the two energy functions based on Wasserstein distance and KL divergence are combined to separately train the context and the extra hyponymy, which can better express the semantic range of a word.
	
	\subsection{Embedding visualization}
	We visualize the trained word embedding data, hoping to map each word into a circle in the plane. The word will be a circle instead of an ellipse when visualizing because we use a spherical covariance matrix to train the word embeddings. We use \emph{principal component analysis} \cite{jolliffe1986principal} to reduce the high-dimensional mean data $\vec{\mu}_{w}$ into two dimensions as the center of the circle represented by each word. The radius of the circle is controlled based on the numerical value of the word covariance matrix $\Sigma_{w}$.  Figure 1 shows the results of the visualization.

	\begin{figure}[htb]
			\center{\includegraphics[width=5cm]  {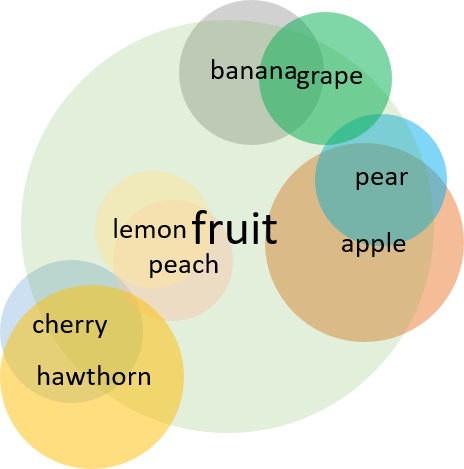}}
	\center{\includegraphics[width=5cm]  {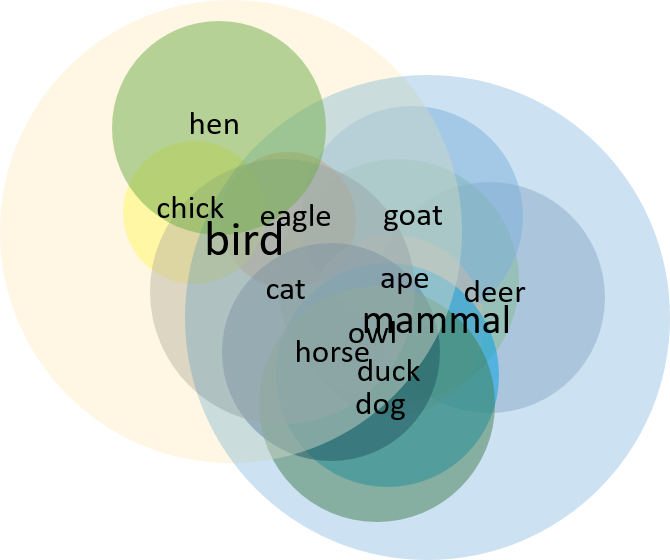}}
		\caption{\label{1} Two examples of visualization results. From the figure we can see that the size of the word circle is a good representation of its semantic coverage.}
	\end{figure}
	
	We find that mapping words into circles on a plane allow us to observe similarity and entailment relationships between words visually.
	
	\subsection{Document classification accuracy}
	Our document classification task uses the word mover distance method \cite{kusner2015word} to test the performance of different word embedding results.
	
	We evaluate on 6 supervised document datasets: BBCSPORT, TWITTER \cite{sanders2011sanders}, RECIPE,  CLASSIC, REUTERS(8 class version described in \cite{cachopo2007improving}) and AMAZON. The experimental results are reported in Table 3.
	
	\begin{table}[h!]
		\centering
		\begin{tabular}{c|c|c|c}
			\toprule
			Dataset & SG & W2G & WDG\\
			\hline
			BBCSPORT & 7.1 & 7.9 & \textbf{6.1} \\
			TWITTER & 30.2 & 30.7 & \textbf{29.9} \\
			RECIPE & 49.9 & \textbf{49.4} & 50.6 \\
			CLASSIC & 4.7 & 5.1 & \textbf{4.5} \\
			REUTERS & 4.6 & 5.5  & \textbf{4.4} \\
			AMAZON & 8.3 & 8.4 & \textbf{7.7} \\
			\midrule
			Avg & 17.5 & 17.8  & \textbf{16.1} \\
			$\Delta$ & - & +0.3 & \textbf{-1.4} \\
			\bottomrule
		\end{tabular}
		\caption{Test error percentage and standard deviation for different text embeddings. The SG, W2G and WDG denote skip-gram \cite{mikolov2013efficient}, Gaussian embedding \cite{vilnis2014word} and Wasserstein Distance Gaussian  model, respectively. We have bolded the best results of the same dataset under different models.
		}
	\end{table}	
	
	We find that the WDG model outperforms the SG and W2G models on five of the six datasets. The average error rate of WDG model is better than that of the other two models.
	
	\section{Discussion}
	\subsection{The meaning of variance}
	The most significant difference between distribution-based and point-based word embeddings is uncertainty. Uncertainty is reflected in variance. We find that WDG model performs better than skip-gram on similarity datasets, which is largely due to variance. It enriches the scope of words and gives a richer representation of words. From another point of view, variance provides the possibility of continuously representing discrete words, and continuation means that data enhancement can be performed by adding tiny perturbations.
	
	The most inherent advantage of distribution is reflected in the judgment of the entailment relationship. When judging the entailment, the training of variance is crucial.  The variance of a particular word of Wasserstein distance energy function tends to the average of the variance of all its context words. KL divergence tends to increase variance, so using KL divergence-based energy function to focus on the extra information of the `IsA' relationship gives the best results in the entailment dataset.
	
	\subsection{The advantage of Wasserstein distance}
	The advantage of Wasserstein distance is that it well describes the distance between two distributions. Even if the support sets of the two distributions do not overlap or overlap very little, the distance between them can still be reflected. Wasserstein distance can solve the problem of predicting non-negative metrics on a finite set. It has recently been successful in image generation\cite{arjovsky2017wasserstein,tolstikhin2017wasserstein}.
	
	In the field of NLP, under the assumption that words are regarded as a multi-dimensional distribution, the advantage of Wasserstein distance can also be utilized to train word embeddings.
	
	\subsection{The extensibility of the model}
	Our model can select different energy functions for different types of information when introducing external information. The experimental results also prove that targeted energy function selection can achieve better results.
	
    We find that our model achieves some advantages over alternative methods in the downstream document classification task. However, the gains are not significant. The reason may be a drawback of non-contextual word embeddings. It only trains against the most basic natures such as similarity and does not fully integrate with downstream tasks. If we can grasp the context information of downstream tasks in real time, the model could be able to achieve better results.
	
	The document classification task does not construct a new RNN or LSTM network based on distribution but only utilizes the distance information in the word embeddings. In the future, if the distribution-based neural network can be constructed, directly using the Gaussian word embedding as the input of the network will make full use of its properties.
	
	\section{Conclusion}
	In this paper, we introduced a Gaussian word embedding model based on Wasserstein distance and a model that introduces external information from ConceptNet based on WDG model. Our model can freely select energy function to train different data in a targeted manner. The model has achieved excellent results in the word similarity datasets and the word entailment dataset. We also tested our model in the downstream document classification task and achieved some advantages over alternative methods.

	\bibliography{sunnysc_gaussian}
	\bibliographystyle{aaai}
	
\end{document}